\documentclass[letterpaper, 10 pt, conference]{ieeeconf}  

\IEEEoverridecommandlockouts                              

\usepackage{graphicx}
\usepackage{times} 
\usepackage{amsmath} 
\usepackage{amssymb}  
\usepackage{cite}
\usepackage{textcomp}

\title{\LARGE \bf
Ground Penetrating Radar-Assisted Multimodal Robot Odometry Using Subsurface Feature Matrix 
}

\author{Haifeng Li$^{1}$, Jiajun Guo$^{1}$, Xuanxin Fan$^{1}$ and Dezhen Song$^{2}$
\thanks{*This work was supported in part by National Science Foundation of China under grant 62373365, and by the Fundamental Research Funds for the Central Universities under grant 3122022PY13.}
\thanks{$^{1}$Haifeng Li, Jiajun Guo and Xuanxin Fan are with the Department of Computer Science, Civil Aviation University of China, Tianjin 300300, China.
        Email: {\tt\small hfli@cauc.edu.cn.}}%
\thanks{$^{2}$Dezhen Song is with the Department of Robotics, Mohamed Bin Zayed University of Artificial Intelligence (MBZUAI), Abu Dhabi, UAE.
        Email: {\tt\small dezhen.song@mbzuai.ac.ae.}%
}
}

\begin{document}

\maketitle
\thispagestyle{empty}
\pagestyle{empty}

\begin{abstract}
Localization of robots using subsurface features observed by ground-penetrating radar (GPR) enhances and adds robustness to common sensor modalities, as subsurface features are less affected by weather, seasons, and surface changes. We introduce an innovative multimodal odometry approach using inputs from GPR, an inertial measurement unit (IMU), and a wheel encoder. To efficiently address GPR signal noise, we introduce an advanced feature representation called the subsurface feature matrix (SFM). The SFM leverages frequency domain data and identifies peaks within radar scans. Additionally, we propose a novel feature matching method that estimates GPR displacement by aligning SFMs. The integrations from these three input sources are consolidated using a factor graph approach to achieve multimodal robot odometry. Our method has been developed and evaluated with the CMU-GPR public dataset, demonstrating improvements in accuracy and robustness with real-time performance in robotic odometry tasks.
\end{abstract}

\section{Introduction}
The task of localizing mobile robots or vehicles without relying on the global positioning system (GPS) becomes particularly difficult in adverse weather conditions like heavy snow, dense fog, or dust storms or season changes, as conventional sensors like cameras or lidars tend to malfunction or unable to find proper correspondence. Nonetheless, the underlying structure of the roads is consistent and generally unaffected by such weather or seasonal changes. Ground penetrating radar (GPR), which can detect subsurface objects, offers a promising alternative for robot localization. Despite this potential, current techniques face challenges in robustness and accuracy due to the unreliable and ambiguous nature of GPR signal registration. 

In this work, we introduce an innovative and efficient subsurface feature matrix (SFM) derived from GPR B-scan imagery, building upon our prior research \cite{Gjjpaper1}, where we identified a subsurface feature known as the dominant energy curve (DEC). Despite its utility, DEC's reliability requires enhancement, prompting the development of the improved SFM technique presented here, which offers greater robustness and compatibility with robotic odometry. The SFM harnesses frequency domain peaks to effectively estimate GPR's relative positions. This advancement enables us to create a sensor fusion strategy for odometry that integrates GPR, inertial measurement unit (IMU), and wheel encoder within a factor graph optimization framework. We have implemented and validated our method using the publicly available CMU-GPR dataset. Our findings indicate that the proposed algorithm significantly enhances accuracy and robustness while maintaining real-time performance in robotic localization tasks. Specifically, the average root mean square error (RMSE) of the GPR component across various scenarios is 2.23 m, and the overall average RMSE for the combined odometry system stands at 0.568 m.

\begin{figure}[t]
      \centering
      \includegraphics[width=0.45\textwidth]{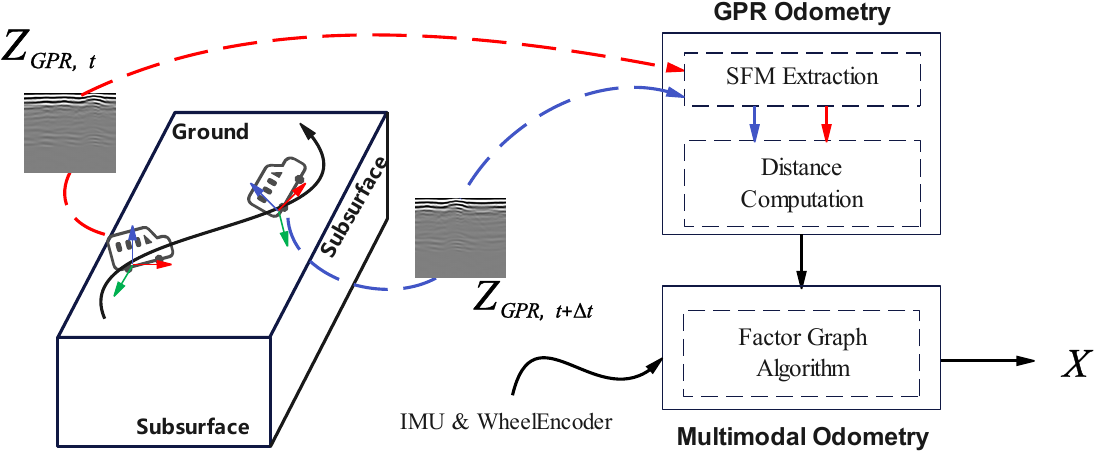}
      \caption{Illustration of our SFM based multimodal odometry algorithm.}
      \label{Fig:firstPic}
\end{figure}

\section{Related Works}

The most relevant related works include GPR usage in various robotic  applications, subsurface reconstruction and mapping, and GPR-based localization and mapping. 

A significant portion of the use of GPR in robotics pertains to nondestructive evaluation tasks, such as inspecting bridge decks\cite{Lim-TASE-2014,Ahmed-sensors-2020}, detecting utilities\cite{Chen-IJCAI-2011,Yang-GPR-2014}, examining runways at airports\cite{Li-RAL-2021,Li-TGRS-2023}, and exploring planetary surfaces\cite{Furgale-FSR-2010,Lai-Nature-2020}. In these scenarios, GPR functions as a sensor to locate and identify particular subsurface objects or features. A common application with a GPR mounted on a robot involves subsurface reconstruction and mapping, where the robot uses a GPR to scan a designated area to create a subsurface map\cite{Kouros-IROS-2018,Chou-ICRA-2018,Chou-TRO-2021}. However, these studies focus on object detection and do not explore the possibility of using the GPR modality for robot localization.

The use of GPR in robotic localization and mapping has recently attracted increased interest from researchers in robotics. The first studies on robot location using GPR are introduced in \cite{Cornick-JFR-2016}, where a Localization GPR (LGPR) module is presented. This module involves a GPR apparatus mounted on a ground vehicle, enabling the vehicle to determine its location using a preexisting map. The approach constructs a grid map that contains a collection of rectangular 3D GPR raw signals with attached location labels. These are prepared ahead of time, allowing new GPR observations to be aligned with the grid map to ascertain the GPR's global position. This method has been advanced to support autonomous navigation in challenging weather conditions using the same LGPR system \cite{Ort-IROS-2020}. A key limitation of the LGPR approach is its reliance on raw GPR data as mapping features, which makes it susceptible to fluctuations in the dielectric properties of the medium during mapping and localization. Consequently, the approach may lack consistency and robustness for prolonged use. In addition, the LGPR method requires initial access to the GPS signal during its operation.

Baikovitz et al. present an innovative GPR-based odometry approach, as described in \cite{Alexander-IROS-2021}, which conceptualizes localization as an inference problem within a factor graph framework. This technique emphasizes directly learning relative sensor models from GPR data, translating nonsequential GPR image pairs into robot motion estimations. They also introduce the CMU-GPR dataset \cite{Alexander-arXiv-2021}, an open-source resource aimed at enhancing research in subsurface-assisted perception for robotic navigation, utilized to validate their approach. However, the effectiveness of this method is highly dependent on the quality of the learned sensor model obtained, which could be affected by environmental conditions and the precision of the manually annotated training datasets. Acquiring comprehensive GPR data for training is challenging both in terms of effort and expense, particularly for real-world implementations.

Therefore, extracting stable features from GPR data to perform mapping and localization is a reasonable choice. However, this task is quite challenging, as GPR data often have significant signal clutters. Skartados et al.\cite{skartados-EUSIPCO-2019} propose a feature-based GPR self-localization method by isolating spatiotemporal salient regions on consecutive GPR traces. The work assumes a simulated GPR model with subsurface utility pipeline priors. However, neither the GPR model nor the requirement for the presence of subsurface pipelines are easily met in general applications.

Our team has extensive expertise in GPR-based robotic inspection and mapping over the years. We have utilized GPR in various applications like airport runway inspections\cite{Li-RAL-2021,Li-TGRS-2023}, utility pipeline rebuilds\cite{Li-TASE-2020}, and subsurface mapping efforts\cite{Chou-TRO-2021}. In our previous study\cite{Gjjpaper1}, we introduced a one-dimensional subsurface feature called DEC for performing localization tasks. In this study, we create a two-dimensional SFM by incorporating an additional dimension from spatially adjacent features, the SFM offers a better contextual and precise representation of the underground environmental conditions. Thanks to this enhancement, SFM offers a much improved feature transform for using GPR in the proposed multimodal robot odometry.

\begin{figure}[t]
      \centering
      \includegraphics[width=0.45\textwidth]{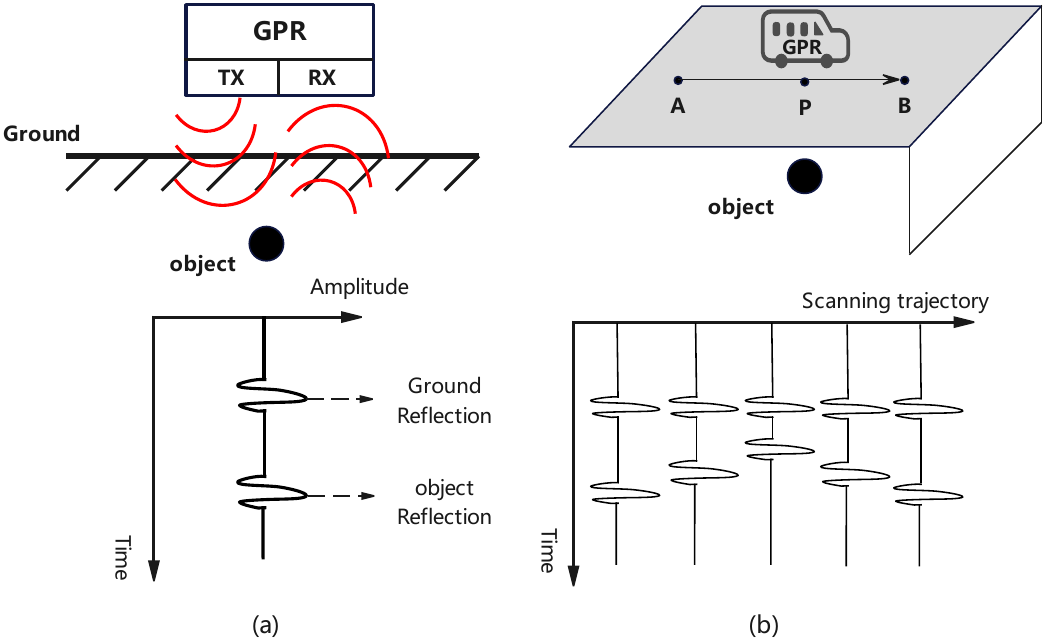}
      \caption{A simplified visualization of GPR working principle from ~\cite{Gjjpaper1}.}
      \label{Fig:GPR-Principle}
\end{figure}

\begin{figure}[t]
      \centering
      \includegraphics[width=0.45\textwidth]{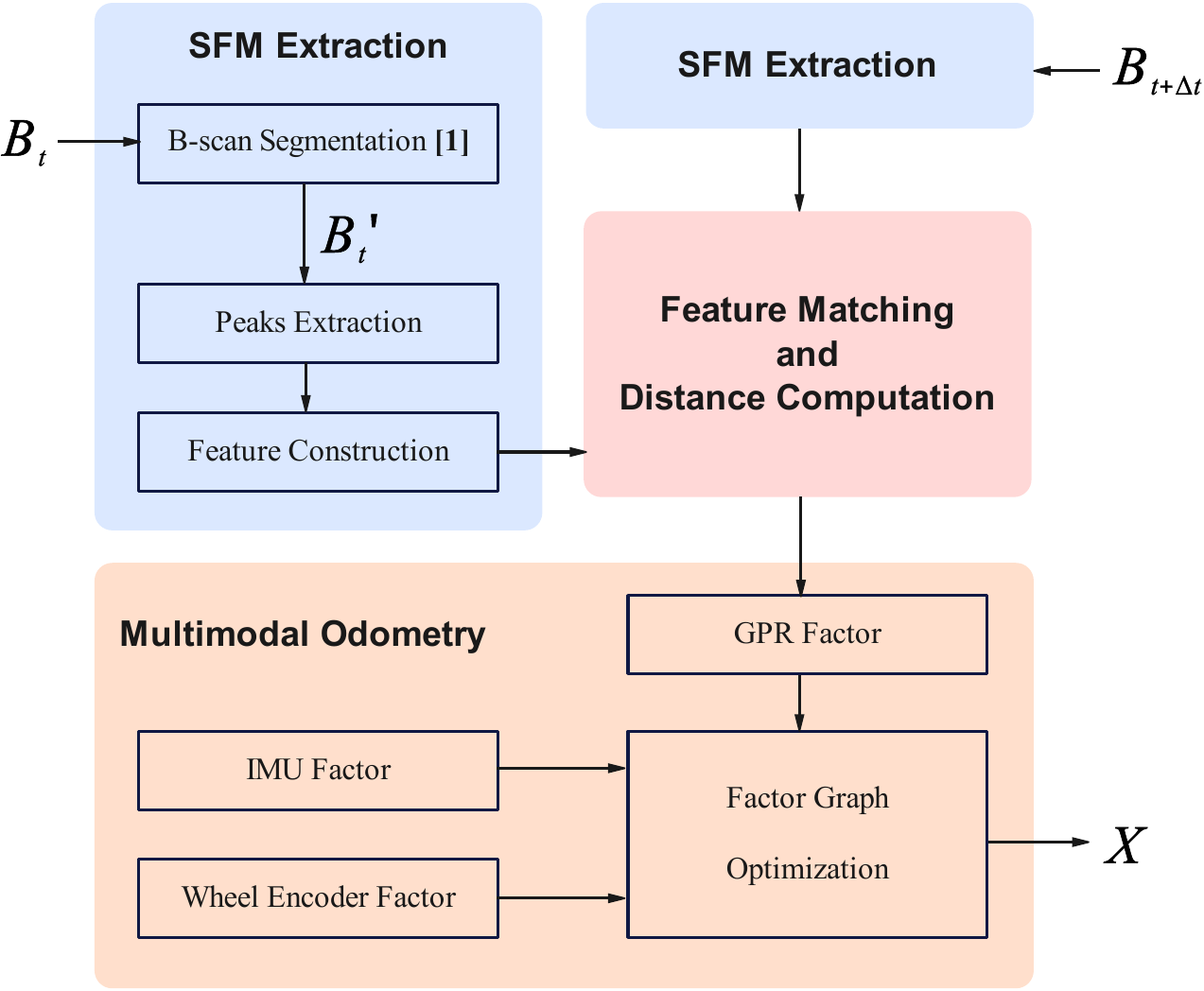}
      \caption{The algorithm flow chart of the paper}
      \label{Fig:algorithm}
\end{figure}

\section{Problem Formulation}

\subsection{Working  Principles of a GPR}

Before presenting the entire algorithm, it is essential to revisit the fundamental principles of a GPR. Fig.~\ref{Fig:GPR-Principle} illustrates the concept of GPR, as elaborated in our earlier study~\cite{Gjjpaper1}. A GPR system is comprised of a transmitter (TX) for sending radar signals and a receiver (RX) for capturing these signals. Reflection occurs when the radio pulse encounters a boundary with changing dielectric permittivity, and the RX receives this reflected pulse. The time taken for radio pulses to travel out and back allows the GPR to create an A-scan, documenting the signal's amplitude against travel time at a specific GPR location (refer to Fig.~\ref{Fig:GPR-Principle}(a)). In Fig.~\ref{Fig:GPR-Principle}(b), an object is situated below point $P$ along the path $AB$ that the GPR traverses. When the GPR reaches point $P$, depicted at the top of Fig.~\ref{Fig:GPR-Principle}(a), the radio pulse from the TX generates two prominent reflections, one resulting from the ground and another from the buried object. The resultant signal is displayed at the bottom of Fig.~\ref{Fig:GPR-Principle}(a), where the upper peak corresponds to the ground and the lower peak to the object. Throughout the journey from $A$ to $B$, the antenna captures a sequence of A-scan signals. A B-scan is a chronological arrangement of these A-scan signals, with a sample depicted at the bottom of Fig.~\ref{Fig:GPR-Principle}(b). The variation in the object’s reflection peaks indicates the changing distance between the GPR and the object.

\subsection{Notations and Problem Definition}
To describe our problem, we define the following notations.

\begin{itemize}

\item $\{W\}$: A 3D right-handed global coordinate system.


\item $\{B\}$: A 2D B-scan coordinate. $X$ and $Y$ axes are the vertical and the horizontal axes, respectively.


\item $\mathbf{Z}$, observations obtained by sensors where  IMU, wheel encoder, and GPR readings are $\mathbf{Z}_{\mbox{\tiny IMU}}$, $\mathbf{Z}_{\mbox{\tiny wheel}}$,  and $\mathbf{Z}_{\mbox{\tiny GPR}}$, respectively.

\item $L$ and $D$ are the maximum width and depth indices of a B-scan image, respectively. Note that $D$ is also the maximum depth index of the A-scans. They can be understood as the width and height of the B-scan image.

\item $A=\{a_{x}|x=1,2,3,\dots,D\}$, A-scan signal obtained by GPR, $x$ is the position/time index, $a_{x}$ is the signal strength value at the index $x$. When there is more than one A-scan, we employ the subscript $y$, e.g., $A_y$ means the $y$-th A-scan.

\item $B_{t}(x,y)=\{a_{x}|a_{x}\in A_{y},\ \mbox{and}\  y=1,2,3,\dots,L\}$, is the $x$-th signal value in the $y$-th A-scan $A_{y}$ at time $t$. Hence, it forms an image indexed by $(x,y)$ at time $t$. $1\leq x \leq D$ is the depth range of each $A_{y}$, and $1\leq y \leq L$ is the range of horizontal position index along the GPR travel trajectory.

\item $S_{t}$, a two-dimensional subsurface feature matrix extracted from $B_{t}(x,y)$ with its width $L_{S}$ and height $D_{S}$ which will be explained later.

\item $\mathbf{x}_{t}=\{\mathbf{p}_{t},\mathbf{v}_{t},\mathbf{R}_{t}\}$, the state of the robot at time $t$, $\mathbf{p}_t$ is the position of the robot in $\{W\}$, $\mathbf{v}_t$ is the velocity, the rotation matrix $\mathbf{R}_t$ describes the orientation of the robot.

\item $u_t$, the travel distance of the robot between time $t-\Delta t$ and $t$ which is a one-dimensional scalar with plus or minus sign as direction.

\item $X=\{\mathbf{x}_{0},\mathbf{x}_{1},\mathbf{x}_{2},\dots,\mathbf{x}_{k}\}$, a set of all the states of the robot, $k$ stands for the time index of the states during the entire localization process.

\end{itemize}

With the notations above, we define our problem as follows.

{\it Definition 1}: 
 Given $\mathbf{Z}_{\mbox{\tiny GPR}}$, $\mathbf{Z}_{\mbox{\tiny IMU}}$ and $\mathbf{Z}_{\mbox{\tiny wheel}}$, determine $X$.

\section{ALGORITHM}
The pipeline of our algorithm is depicted in Fig.~\ref{Fig:algorithm}. The methodology is divided into three steps. The initial step, SFM Extraction, generates an SFM from each B-scan image. This part consists of three subcomponents: Details Retention aims to mitigate clutter effects; Peaks Extraction identifies peaks in the signal from processed B-scan images and employs these peaks to build the SFM in the Feature Construction subsection. Once the SFMs are obtained, optimization-based techniques are applied to determine the similarity between two SFMs and calculate the robot's travel distance from this similarity. Finally, we employ a factor graph optimization algorithm to integrate the GPR module, IMU, and wheel encoder module to accurately reconstruct the robot's trajectory.

\begin{figure}[t]
      \centering
      \includegraphics[width=0.45\textwidth]{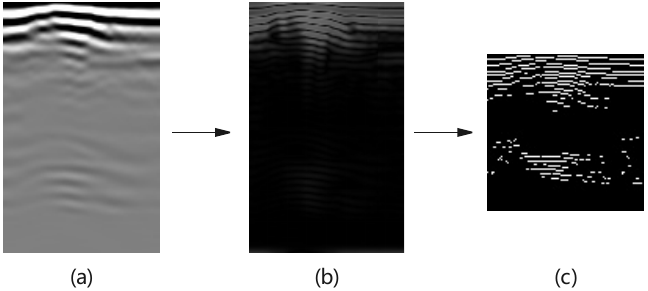}
      \caption{Sample $B_t(x,y)$ in (a), $B'_t(x,y)$ in (b),  and $S_{t}$ in (c)}
      \label{Fig:middle}
\end{figure}

\subsection{SFM Extraction}

\subsubsection{Details Retention}
During the GPR scanning process, substantial reflection signals might be obscured by various background noises. Therefore, it is advisable to minimize noise and irrelevant reflections to better highlight significant reflection signals. This section builds on our earlier research~\cite{Gjjpaper1}. The GPR data $Z_{\mbox{\tiny GPR}}$ can be described as a sequence of B-scan images captured over time. For each B-scan image $B_{t}(x,y)$ from $Z_{\mbox{\tiny GPR}}$, utilize the algorithm proposed in~\cite{Gjjpaper1} to achieve B-scan Segmentation as part of the Feature Extraction phase. Following Frequency Domain Decomposition, alongside Detail preservation and Restoration procedures, we transform $B_{t}(x,y)$ into $B_{t}^{'}(x,y)$. Fig.~\ref{Fig:middle}(b) provides a sample outcome. The new B-scan image $B_{t}^{'}(x,y)$ consists of a set of new A-scans $A^{'}_{y}(x)$, which preserve peaks while dampening low amplitude, low-frequency variations,
\begin{equation}
    A_{y}'(x) = B_{t}'(x,y), ~\forall x,y.
\end{equation}
With the detail retention phase completed, we proceed to extract peaks from $B_{t}'(x,y)$.

\subsubsection{Peaks Extraction}
In each processed A-scan signal $A'_{y}$, the key peaks mark the locations where substantial reflections occur as electromagnetic waves interact with objects or boundaries exhibiting changes in dielectric constant. These peaks convey crucial data on both the two-way travel time of electromagnetic waves and the strength of the reflections. To extract this information from these peaks, we utilize the approach detailed in \cite{Li-TASE-2020}. This study describes the application of a damped sinusoidal model to fit the processed reflected signal $A'_{y} \in B_{t}'$. For each filtered pulse response, the expression is given by
\begin{equation}
    A^{'}_{t}=\beta_{t}e^{-\alpha_{t}i}\cos(\omega_{t}i+\phi_{t})+\gamma_{t}
\end{equation}
where $A^{'}_{t}$ represents the signal captured by GPR at the time $t$ index, $\beta_{t}$ indicates the amplitude, $\omega_{t}$ is the angular frequency, $\phi_{t}$ is the phase shift, $\gamma_{t}$ denotes the model's offset, and $\alpha_{t}$ is the attenuation constant determined by the material's properties.

Then we can estimate the peak positions by solving the optimization problem as follows,
\begin{equation}
\begin{aligned}  \underset{\hat{\alpha}_{i},\hat{\beta}_{i},\hat{\gamma_{i}},\hat{\omega_{i}},\hat{\phi_{i}}}{\arg \min}\sum_{i}||\hat{A}^{'}_{t}-A^{'}_{t}||^{2}_{\sum} \\
   s.t. \ \hat{A}^{'}_{t}=\hat{\beta_{t}}e^{-\hat{\alpha_{t}}i}\cos(\hat{\omega_{t}}i+\hat{\phi}_{t})+\hat{\gamma}_{t}
\end{aligned}
\end{equation}
where symbol hat $\hat{}$ indicates the estimators and $||\cdot||_{\sum}$ denotes the Mahalanobis distance. We employ Maximum Likelihood Estimate (MLE) to solve this problem, in other words, we use known A-scan signals sample information data to deduce the parameter values in the damped sinusoid model. At the end of this step, we obtain the time and amplitude of each peak.

We now proceed to determine the travel times of the signal at each peak. By using the speed at which electromagnetic waves propagate, these travel times can be converted into travel distances. Utilizing the GPR's specific parameters, these distances can then be converted into pixel indices in $\{B\}$. Apply these calculations to all A-scans $A'_{y} \in B'_{t}$. Consequently, define a single peak's position as $e_{i}=[r_{i},d_{i}]^{\mathsf{T}}$, which represents the $i$-th peak location derived from $B'_{t}(x,y)$, where $r_{i}$ and $d_{i}$ are the horizontal and vertical coordinates in $\{B\}$. Define the set of peaks $M$ as:
\begin{equation}
    M_{t}=\{e_{i}|i=1,2,3,\dots,m\}
\end{equation}
where $m$ denotes the number of peaks in $B'_{t}(x,y)$.

\subsubsection{Feature Construction}
The set of peaks $M_{t}$ serves as an efficient observation from GPR at moment $t$. The locations and magnitudes of these peaks provide insights into the overall subsurface environmental conditions.

Let us transform the data in $M_{t}$ into a two-dimensional format, SFM $S_{t}$, following a two-step process. Initially, for each $e_{i}$ in $M_{t}$, determine its position within $S_{t}$. The coordinates of $e_{i}$ in $S_{t}$ are given by:
$ (\lfloor \frac{d_{i}D_{S}}{D} \rfloor ,\lfloor \frac{r_{i}L_{S}}{L} \rfloor ) $
where $D_{S}$ and $L_{S}$ are the dimensions of $S_{t}$, while $D$ and $L$ denote the B-scan $B_{t}(x,y)$ dimensions. Next, convert the grayscale amplitude $m_{i}$ of $e_{i}$ to the nearest level using $\lfloor \frac{10 m_{i}}{255} \rfloor \in [0,10]\cap \mathbb{Z}$, resulting in one of 11 possible discrete levels. All other matrix values are assigned to zero to complete $S_{t}$.

Fig.~\ref{Fig:middle}(c) presents a sample $S_t$ depicted in black and white. This sample $S_{t}$ reduces clutter in B-scan images, thereby emphasizing reflection signals and depicting the key reflection points of electromagnetic waves at particular positions and depths on the surface. The SFM effectively illustrates the spatial distribution and intensity of electromagnetic waves that are well-reflected below the ground, providing an equivalent representation of underground environmental information. Additionally, the benefit of creating a matrix lies in facilitating future feature matching computations, as will be demonstrated next.

\subsection{Feature Matching and Distance Computation}
If SFMs $S_{t}$ and $S_{t+\Delta t}$ obtained at time $t$ and $t+\Delta t$ are spatially correlated, it is highly likely that they possess similar components which can be exploited for feature matching purposes.  To evaluate the similarity between $S_{t}$ and $S_{t+\Delta t}$, we adopt the cosine distance as a metric. The cosine distance between two matrices is defined as,
\begin{equation}
    C(A,B)=1-\frac{\sum_{i=1}^{m \times w}A_{i} \cdot B_{i}}{\sum A^{2}_{i}\sum B^{2}_{i}}
\end{equation}
where $A$ and $B$ are two matrices need to compared, and $m$ and $w$ are the width and height of the matrix, respectively.

Between times $t$ and $t+\Delta t$, the robot moves, and therefore $S_{t}$ and $S_{t+\Delta t}$ are not identical due to different sensing locations. However, the best match can be found by searching for the horizontal shift in SFM $l(S_{t},S_{t+\Delta t})$. This leads to an optimization problem that estimates $l(S_{t},S_{t+\Delta t})$ by finding the best matching between $S_{t}$ and $S_{t+\Delta t}$,
\begin{equation}
\begin{aligned}
    l^*(S_{t},S_{t+\Delta t})&=\arg\underset{l}{\min} \ C(S_{t}[1,j_{t},D_{S},w], \\ 
    &S_{t+\Delta t}[1,j_{t+\Delta t},D_{S},w]) \\
    s.t. \ &w=L_{S}-|l| \\
    &j_{t}=\max(1,-l+1) \\
    &j_{t+\Delta t}=\max(1,l+1) \\
    &0<|l|<L_S
\end{aligned}
\end{equation}
where $l^*$ indicate optimal solution of the horizontal shift, and $S[1,j,D_{S},w]$ represents for a sub-matrix of $S$ maintaining all rows but with columns from $j$ to $w$. If this optimization problem has a reasonable solution, then $S_{t}$ and $S_{t+\Delta t}$ are spatially correlated.  In fact, the horizontal shift $l^*$ in $\{B\}$ is caused by the longitudinal movement of the robot.  Since a GPR has a fixed sampling frequency and sampling interval between adjacent radar scans $\Delta t$ is short, the robot longitudinal movement or travel distance $u_{t+\Delta t}$ must be proportional to the length of $l^*$,    
\begin{equation}
    u_{t+\Delta t}=K|l^*|,
\end{equation}
where $K$ is the coefficient and $u$ is a one dimensional scalar measured in meters. Applying the above method to the data in $Z_{\mbox{\tiny GPR}}$ and collecting all $u$, we obtain the GPR measurement set $\mathbf{u}$.

\subsection{GPR-Assisted Multimodal Odometry}\label{ssc:fusion}
Using the robot's travel distance derived from GPR measurements, we employ the factor graph\cite{FactorBook} optimization approach to integrate observation data from the IMU and wheel encoder. Within the factor graph optimization algorithm, the IMU offers measurements of acceleration and angular velocity, while the wheel encoder and GPR supply data on the travel distance. To incorporate these observation data into the factor graph, residual construction is necessary.

\subsubsection{IMU Pre-integration Factor}

As for the residual function of the IMU model in the factor graph, we use pre-integration method to construct it. Unlike traditional IMU motion integration, pre-integration  accumulates IMU measurement data over a period of time, establish pre-integration measurement, and ensure that the measurement values are independent of state variable. According to\cite{onManifoldIMU01,onManifoldIMU02}, the residual can be defined as the sum of the difference of orientation, velocity, position and bias obtained by the IMU sensor in $Z_{\mbox{\tiny IMU}}$ as follows, 
\begin{equation}
    ||\textbf{r}_{\mbox{\tiny imu}}(\mathbf{x}_{t-1}, \mathbf{x}_{t})||^2_{\sum_{\mbox{\tiny imu}}} = ||\textbf{r}_{I_{ij}}||^2 + ||\textbf{r}_{b_{ij}}||^2,
\end{equation}
where the left hand side can be viewed as the residual of the IMU factor, and time indices $i$ to $j$ describe the time duration of pre-integration. More specifically, $i=t-1$ and $j=t$. $\textbf{r}_{I_{ij}}$ and $\textbf{r}_{b_{ij}}$ are the pre-integration error residual and bias term estimation errors. More details can be found in  \cite{onManifoldIMU01,onManifoldIMU02}.

\begin{figure}[h]
      \centering
      \includegraphics[width=0.45\textwidth]{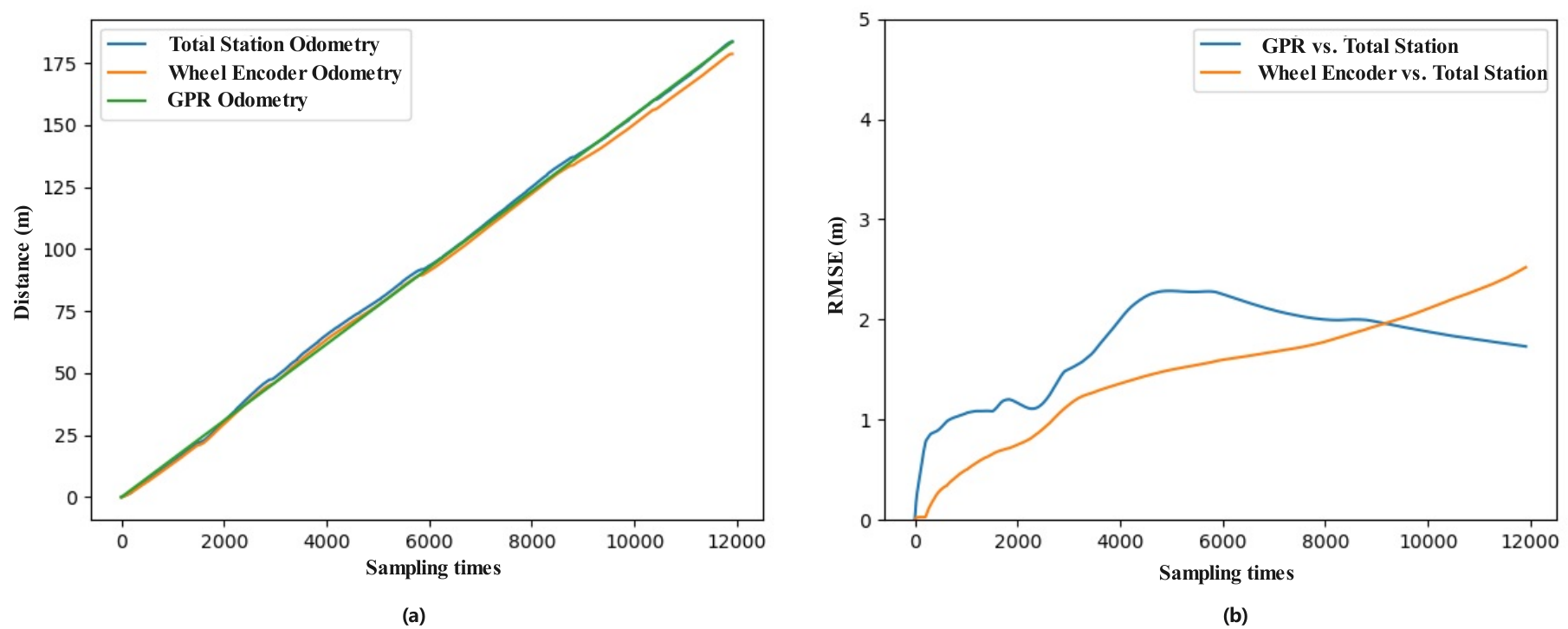}
      \caption{Efficacy of the GPR odometry with SFM: (a) trajectory estimated from individual modalities including total station (ground truth), wheel encoder and GPR. (b) RMSE comparison between the GPR and the wheel encoder modalities.}
      \label{Fig:MotionDisEva}
\end{figure}

\begin{figure}[h]
      \centering
      \includegraphics[width=0.40\textwidth]{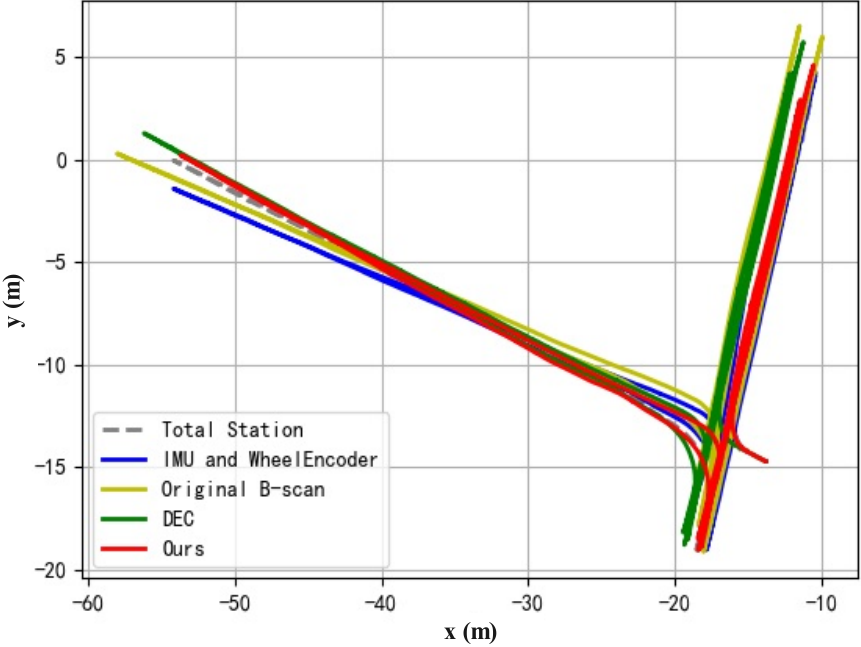}
      \caption{Trajectory comparison using the Gates$\_$g scene. } 
      \label{Fig:trajs}
\end{figure}

\begin{figure*}[h]
      \centering
      \includegraphics[width=\textwidth]{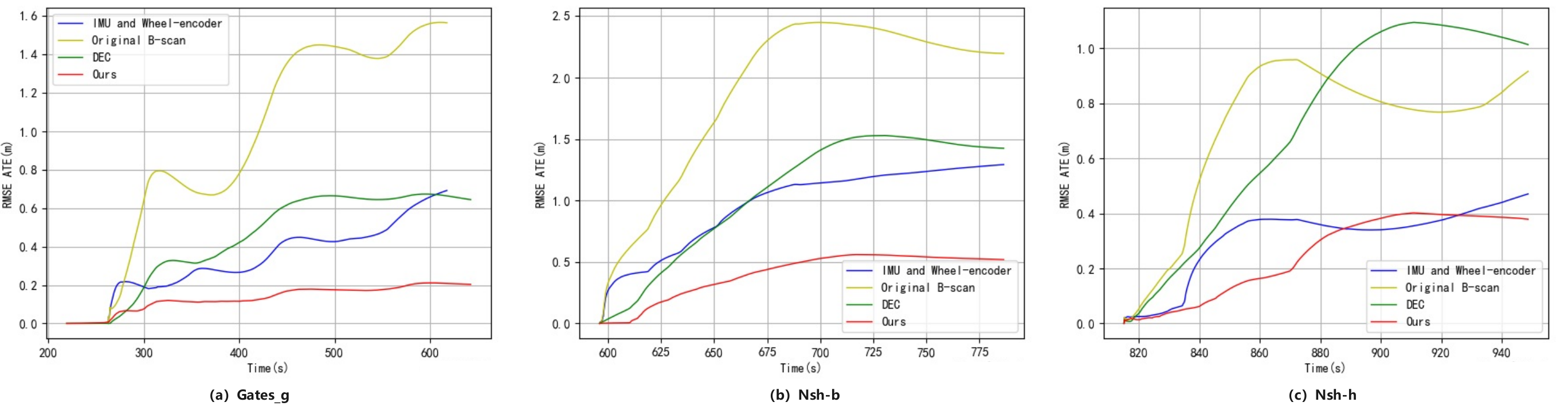}
      \caption{A comparison of RMSE values of the trajectories vs. time under the three datasets.}
      \label{Fig:rmse}
\end{figure*}

\subsubsection{GPR Factor}
The GPR modality provides
observations of the travel distance. The residual has a quadratic form which is defined as the Mahalanobis distance between the travel distance obtained from GPR data and the Euclidean distance from the two adjacent position status, written as,
\begin{equation}
    ||\textbf{r}_{\mbox{\tiny GPR}}(\mathbf{x}_{t-1}, \mathbf{x}_{t})||^2_{\sum_{\mbox{\tiny GPR}}} = ||D(\mathbf{x}_{t-1},\mathbf{x}_{t})\ominus u_{t-1,t}||^2_{\sum_{\mbox{\tiny GPR}}},
\end{equation}
where the function $D(\cdot)$ calculates the Euclidean distance between states $x_{t-1}$ and $x_{t}$, and $u_{t-1,t}$ is the travel distance obtained from the previous SFM matching. 


\subsubsection{Wheel Encoder Factor}
The wheel encoder factor shares the similar format with the GPR factor which can also be treated as a distance observation,
\begin{equation}
    ||\textbf{r}_{\mbox{\tiny wheel}}(\mathbf{x}_{t-1},\mathbf{x}_{t})||^2_{\sum_{\mbox{\tiny wheel}}} = ||D(\mathbf{x}_{t-1},\mathbf{x}_{t})\ominus z^{\mbox{\tiny wheel}}_{t-1,t}||^2_{\sum_{\mbox{\tiny wheel}}},
\end{equation}
where $z^{\mbox{\tiny wheel}}_{t-1,t}$ is the measurement data obtained from wheel encoder $Z_{\mbox{\tiny wheel}}$.

\subsubsection{Loss Function}
With the above sensor residuals, the entire loss function of the problem using a factor graph is as follows,
\begin{align}
\nonumber    X = \arg \underset{X}{\min} \sum^k_{t=1}\{&||\textbf{r}_{\mbox{\tiny imu}}(\mathbf{x}_{t-1}, \mathbf{x}_{t})||^2_{\sum_{\mbox{\tiny imu}}}+ \\ 
\nonumber    &||\textbf{r}_{\mbox{\tiny GPR}}(\mathbf{x}_{t-1}, \mathbf{x}_{t})||^2_{\sum_{\mbox{\tiny GPR}}} + \\ 
    &||\textbf{r}_{\mbox{\tiny wheel}}(\mathbf{x}_{t-1},\mathbf{x}_{t})||^2_{\sum_{\mbox{\tiny wheel}}}\},
\end{align}
where $k$ is the state time index. Finally, we obtain the updated robot status set $X$.

\section{Experiments}
Our multimodal sensor fusion odometry is assessed using the publicly available CMU-GPR Dataset. Initially, we provide a brief overview of the dataset. Subsequently, we examine the efficacy of the travel distance evaluation algorithm, relying solely on the GPR modality derived from subsurface characteristics. Finally, we conduct both qualitative and quantitative assessments of the multimodal sensor fusion-based odometry, contrasting it against multiple alternatives.

\subsection{Dataset Introduction}
Our study utilizes the CMU-GPR dataset \cite{Alexander-arXiv-2021}, an open-source collection specifically designed for research in robotic navigation aided by subsurface data. This dataset comprises multiple trajectory sequences across three GPS-denied environments: 7 sequences in a basement (\textit{nsh\_b}), 5 on a factory floor (\textit{nsh\_h}), and 3 in a parking garage (\textit{gates\_g}). Collectively, the total trajectory lengths for the \textit{gates\_g}, \textit{nsh\_b}, and \textit{nsh\_h} datasets are 365 meters, 264 meters, and 90 meters, respectively. Each trajectory provides data from a single-channel GPR, a camera, a wheel encoder, and total station sensors. Due to its high accuracy, total station outputs are used as the ground truth in the rest of the experiments.

\subsection{Efficacy of GPR-based Odometry using SFM}

To assess whether SFM feature matching alone is a valid approach for distance measurement, we solely employ the GPR modality in odometry and contrast its performance with that of the total station (considered as the ground truth) and the wheel encoder. Fig.~\ref{Fig:MotionDisEva}(a) presents this comparison. The sub dataset used in Fig.~\ref{Fig:MotionDisEva} is from \textit{gates\_g} environment. The figure clearly shows that using only the GPR modality results in effective odometry outcomes, surpassing the accuracy of the wheel encoder. Fig.~\ref{Fig:MotionDisEva}(b) depicts the RMSE trends over various sampling intervals for both the GPR and wheel encoder. It is evident that the RMSE for wheel encoder-based odometry escalates more rapidly over time compared to that of the GPR. On average, the GPR RMSE is 1.73 m, while the wheel encoder's RMSE is 2.52 m. This demonstrates that GPR SFM is a viable option for odometry, matching the performance of wheel-based methods. Given their complementary nature, it makes sense to integrate them, as we will demonstrate subsequently.

\subsection{Multimodal Odometry}
Next we test the sensor fusion approach presented in Sec.~\ref{ssc:fusion} and compare the overall odometry results with the following counterparts,
\begin{itemize}
    \item \underline{Total station}: the trajectory is obtained from the total station measurement. It is used as ground truth because of its high accuracy.
    \item \underline{IMU and wheel encoder}: This setup combines with IMU and wheel encoder measurement data in odometry. This setup provides a baseline for showing the effectiveness of the GPR modality.
    \item \underline{Original B-scan GPR}: In this setup, we directly used original B-scan images with cosine similarity to calculate the travel distance of the robot. Then the result is fused with IMU and wheel encoder measurement data using the factor graph framework the same way. This setup is to verify the effectiveness of SFM.
    \item \underline{DEC}: To compare with our previous work\cite{Gjjpaper1}, we design multimodal odometry based on DEC instead of SFM.
    \item \underline{CMU learned model}: This represents the state-of-the-art research from Carnegie Mellon University \cite{Alexander-IROS-2021}. 
\end{itemize}
Each odometry method is assessed using the same CMU dataset. Fig.~\ref{Fig:trajs} shows the method trajectories, while Fig.~\ref{Fig:rmse} presents the RMSE comparison across  all three test scenes. The findings indicate that our GPR-assisted multimodal odometry achieves the highest accuracy among all compared methods. Conversely, the Original B-scan and DEC models show the least accurate results, as anticipated. The poor performance of the Original B-scan model is due to signal cluttering impacting accuracy, and the DEC model struggles because its one-dimensional features are not directly linked to displacement estimation.

We also have evaluated our odometry results against those of the CMU Learned model. While their paper does not include trajectory data needed for a point-by-point comparison with other approaches, we are still able to compare trajectory accuracy statistically in the final analysis. The average RMSE of our algorithm in three different environments are 0.468 m, 0.734 m, and 0.439 m. After weighting the trajectories by length, our algorithm's overall RMSE is 0.568 m, compared to the CMU Learned model's 0.59 m. These results indicate our model outperforms the CMU Learned model. Additionally, our method eliminates the need for a training phase, which greatly simplifies deployment and enhances robustness.

\section{CONCLUSION AND FUTURE WORK}
We introduced a GPR assisted multimodal odometry algorithm based on a subsurface feature matrix using a single channel GPR, IMU, and wheel encoder. A novel and more efficient subsurface feature matrix was designed to aid in robot odometry, enabling measurement of the distance traveled by feature matching. Subsequently, we integrated IMU and wheel encoder data with our GPR readings using a factor graph approach to compute the robot's travel path. Our experimental findings demonstrated that the GPR localization method based on multimodal sensor fusion was most effective when evaluated on the public CMU-GPR dataset. Furthermore, our feature extraction algorithm improves the interpretability of the extracted features, ensuring that our odometry, which relies on subsurface features, exhibits greater robustness in various environments.

In the future, we plan to implement our algorithms on a physical robot rather than restricting them to publicly available datasets. Additionally, it is crucial to create more efficient algorithms to decrease computational time.

\bibliographystyle{IEEEtran}

\bibliography{lihf}

\end{document}